\pdfoutput=1
\documentclass[11pt]{article}

\usepackage{acl}

\usepackage{times}
\usepackage{latexsym}
\usepackage{amsmath}
\usepackage{booktabs}
\usepackage{graphicx}
\usepackage{multirow}
\usepackage{bm}

\usepackage{algorithm}
\usepackage{algorithmic}
\usepackage{subfigure}
\usepackage{physics}
\newlength{\lsuperstar}
\usepackage[T1]{fontenc}
\usepackage[utf8]{inputenc}

\usepackage{microtype}

\title{From Mimicking to Integrating:\\ 
Knowledge Integration for Pre-Trained Language Models}
\author{Lei Li\textsuperscript{1}, Yankai Lin\textsuperscript{2,3}, Xuancheng Ren\textsuperscript{1}, 
\textbf{Guangxiang Zhao\textsuperscript{1}, Peng Li\textsuperscript{4}, Jie Zhou\textsuperscript{5}, Xu Sun\textsuperscript{1}} \\
   \textsuperscript{1}MOE Key Lab of Computational Linguistics, School of Computer Science, Peking University \\
     \textsuperscript{2}Gaoling School of Artificial Intelligence, Renmin University of China, Beijing, China
\\
\textsuperscript{3}Beijing Key Laboratory of Big Data Management and Analysis Methods , Beijing, China \\ 
    \textsuperscript{4}Institute for AI Industry Research~(AIR), Tsinghua University, China\\
  \textsuperscript{5}Pattern Recognition Center, WeChat AI, Tencent Inc., China\\
    \texttt{lilei@stu.pku.edu.cn}\quad
    \texttt{xusun@pku.edu.cn}
  }

\begin{document}
\maketitle
\begin{abstract}
% Pre-trained language models~(PLMs) have become the \emph{de facto} standard approach for text classification.
% As many fine-tuned pre-trained language models~(PLMs) with promising performance are generously released, investigating better ways to reuse these models is vital as it can greatly reduce the retraining computational cost and the potential environmental side-effects.
% Since the labeled data for training these models are usually inaccessible due to various reasons such as data privacy, how 
% To effectively reuse these released PLMs,

% 在 abstract 里忽略掉难点 因为篇幅似乎太长了
% The lack of annotated data makes estimating the golden supervision on the union label set for guiding the student intractable.
Investigating better ways to reuse the released pre-trained language models~(PLMs) can significantly reduce the computational cost and the potential environmental side-effects.
This paper explores a novel PLM reuse paradigm, Knowledge Integration~(KI). Without human annotations available, KI aims to merge the knowledge from different teacher-PLMs, each of which specializes in a different classification problem, into a versatile student model.
To achieve this, we first derive the correlation between virtual golden supervision and teacher predictions. We then design a Model Uncertainty--aware Knowledge Integration~(MUKI) framework to recover the golden supervision for the student. Specifically, MUKI adopts Monte-Carlo Dropout to estimate model uncertainty for the supervision integration.
An instance-wise re-weighting mechanism based on the margin of uncertainty scores is further incorporated, to deal with the potential conflicting supervision from teachers.
Experimental results demonstrate that MUKI achieves substantial improvements over baselines on benchmark datasets.
Further analysis shows that MUKI can generalize well for merging teacher models with heterogeneous architectures, and even teachers major in cross-lingual datasets.\footnote{Our code is available at \url{https://github.com/lancopku/MUKI}. Part of the work was done while Yankai Lin and Peng Li were working at Tencent.}
% \footnote{}. Part of the work was done while Peng Li was working at Tencent.}
% , even obtaining close performance with the method trained with labeled data
\end{abstract}

\section{Introduction}

% Text classification tasks have been witnessing remarkable progress 
Large-scale pre-trained language models~(PLMs), such as BERT~\citep{devlin2019bert}, RoBERTa~\citep{Liu2019RoBERTa} and T5~\citep{raffel20t5} have recently achieved promising results after fine-tuning on various natural language processing~(NLP) tasks.
Many fine-tuned PLMs are generously released for facilitating researches and deployments.
Reusing these PLMs can greatly reduce the computational cost of retraining the PLM from scratch and alleviate the potential environmental side-effects like carbon footprints~\citep{strubell-etal-2019-energy}, thus making NLP systems greener~\citep{schwartz2020green}.
A commonly adopted model reuse paradigm is knowledge distillation~\citep{Hinton2015Distilling,romero15fitnet}, 
where a student model learns to mimic a teacher model by aligning its outputs to that of the teacher.
% which transfers learned knowledge from a teacher model to a student model via matching the outputs between the teacher model and the student. 
In this way, though achieving promising results with PLMs~\citep{Sun2019PatientKD,Jiao2019TinyBERT}, the student is restricted to perform the same task as the teacher model, thus restricting re-utilization of abundant available PLMs fine-tuned on different tasks, e.g., models fine-tuned on various label sets or even different datasets.
%%% Figure to illustrate Knowledge Amalgamation
\begin{figure}[t!]
    \centering
    \includegraphics[width=0.95\linewidth]{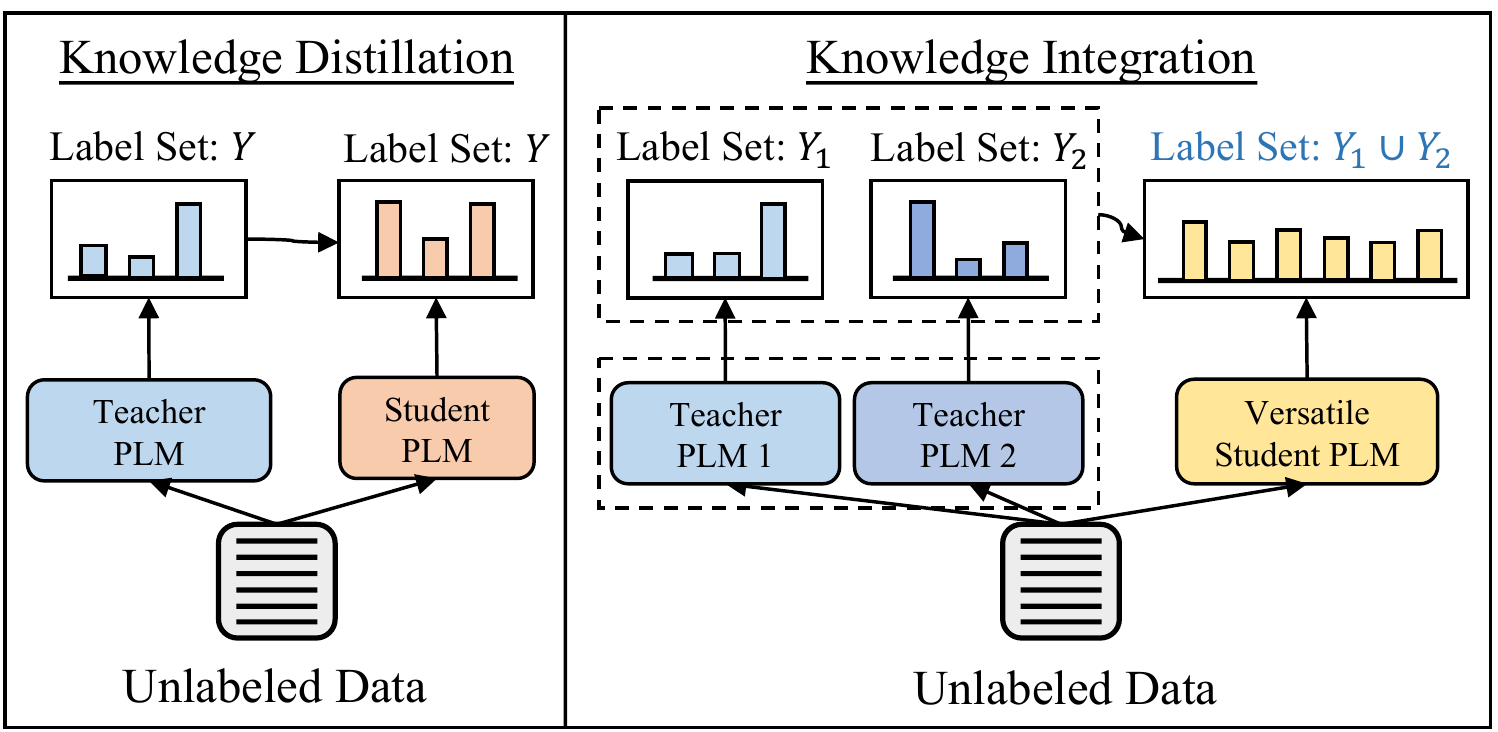}
    \caption{Comparison of knowledge distillation~(KD) and knowledge integration~(KI). KD assumes that the student performs predictions on the identical label set with the teacher, while KI trains a student model that is capable of performing classification over the union label set of teacher models.}
    \label{fig:ka_tc}
\end{figure}
%%% Figure to illustrate KD & KA 

% emphasize the key different here 
In this paper, we generalize the idea of KD from mimicking teachers to integrating knowledge from teachers, and propose \emph{Knowledge Integration}~(KI) for PLMs.
Given multiple fine-tuned teacher-PLMs, each of which is capable of performing classification over a unique label set, KI aims to train a versatile student that can make predictions over the union of teacher label sets.
As the labeled data for training the teachers may not be publicly released due to data privacy issues, we assume no human annotations are available during KI.
% , which makes KI more practical in real-world applications
The benefits of KI are two-fold. 
First, compared to KD, KI can make full use of the released PLMs specializing different tasks.
Besides, the ability of the versatile student, i.e., the label set coverage, can be improved over time by integrating newly released teacher models.
Figure~\ref{fig:ka_tc} illustrates the main difference between KD and KI.

% which is infeasible in many scenarios as the annotated data may not be publicly available due to a series of reasons such as privacy issues.
%Besides, deploying PLM classifiers separately for different text class sets is challenging due to the huge memory demands.
% proportional to the number of PLMs.
% \textbf{ in a one-to-one style}
% , thus motivating us to explore a better all-in-one model reuse framework for merging knowledge from different PLMs.
% Although PLMs provide a concrete foundation for text classification, fine-tuning and tailoring the PLMs require intensive computation and abundant annotated data, which are hardly accessible and available.
% To tackle the difficulty in applying PLMs to diverse classification tasks, 

% To promote better utilization of PLMs specializing on different classification tasks, 

As no annotations are available, the core challenge of KI lies in the integration of outputs from teachers to form golden supervision, i.e., the class probability distribution over the union label set, for guiding the student. 
Through theoretical derivation, we first build the bridge between the teacher predictions and the golden supervision, which indicates that the key to recovering such supervision is to identify the adequate teacher for each instance.
However, due to the over-confident problem of PLMs~\citep{desai2020calibration}, selecting qualified teachers for unlabeled instances is non-trivial, and our exploration shows that prediction entropy is misleading. 
Inspired by Monte-Carlo Dropout~\citep{gal2016dropout}, we inject parameter perturbations to the teacher models during inference and then estimate the model uncertainties over averaged predictions for indicating the possible correct teacher model.
% , which explores the model uncertainty to estimate golden supervision over the union label set for guiding the student.
% Intuitively, the model predictions are consistently confident for instances in the model specialty even with minor perturbations on the model parameters. 
% On the contrary, instances with categories out of the teacher specialty will be predicted with more random results under perturbations, as they are not well fitted by the model parameter.
% Therefore, the uncertainty of averaged predictions can be utilized for identifying the correct teacher, which is compatible with recent progress in Bayesian neural networks~\citep{gal2016dropout,uncertainty17kendall}.
% We adopt Monte-Carlo Dropout~\citep{gal2016dropout} to accurately estimate the model uncertainty.
Our Model Uncertainty--aware Knowledge Integration~(MUKI) framework is then proposed based on the estimated model uncertainty.
Specifically, the golden supervision is approximated by either taking the outputs of the most confident teacher, or softly integrating different teacher predictions according to the relative importance of each teacher.
Furthermore, for instances on which teachers achieve close uncertainty scores, we introduce a re-weighting mechanism based on the margin of uncertainty scores, to down-weight the contribution of instances with potential conflicting supervision signals.
% As MUKA takes the potential supervision conflict into consideration and only operates on the prediction level, it can generalizes well for amalgamating heterogeneous PLMs specializing different tasks.

Experimental results show that MUKI can successfully achieve the goal of knowledge integration, significantly outperforming baseline methods, and even obtaining comparable results with models trained with labeled data.
% , and generalize well to challenging settings, including merging knowledge from multiple teacher models, heterogeneous teachers with different architectures, or cross-dataset even cross-lingual teacher models.
Further analysis shows that MUKI can produce supervision close to the golden one and generalize well for merging knowledge from heterogeneous teachers with different architectures, or even cross-lingual teacher models. 
The main contributions of this work can be
summarized as follows:
(1) We explore knowledge integration for PLMs, which is capable of making full use of released PLMs with different label sets and has great extendability.
(2) We present MUKI, a generalizable KI framework, which integrates the knowledge from teachers according to model uncertainty estimated via Monte-Carlo Dropout and re-weights the instance contribution based on the uncertainty margin.
(3) Experimental results demonstrate that MUKI is effective and generalizable, significantly outperforming baselines.

% provides accurate supervision for guiding the student, even achieving closing performance to the supervised upper bound on benchmark datasets. Further analysis shows that MUKI is generalizable and extendable for amalgamating heterogeneous teachers and cross-dataset teachers.
% for merging knowledge from various teachers and generalizes well for amalgamating heterogeneous and cross-dataset teachers.

\section{Knowledge Integration for PLMs}
In this section, we first give the task formulation for knowledge integration, followed by the elaboration on the proposed MUKI framework.
% , which consists of supervision estimation and instance re-weighting mechanism based on model uncertainty
% Figure~\ref{fig:uka_framework} gives an overview of our proposal. 
\subsection{Problem Formulation}
% We formulate knowledge integration for text classification with PLMs as below.
Given $N$ teacher PLM models $TS = \left\{T_1, \dots, T_N \right\}$, where each teacher $T_i$ specializes in a specific classification problem, i.e., a set of classes $Y_i$, knowledge integration aims to train student model $S$ for performing predictions over the comprehensive class set $Y = \bigcup_{i=1}^{N} Y_{i}$, with an unlabeled dataset $\mathcal{D}$.
% As previous studies~\citep{Vongkulbhisal19UHC,Thadajarassiri21SKA} show that merging teachers with overlapping classes is easy, 
We assume that for each instance in $\mathcal{D}$ there is at least one teacher capable of handling it and we focus on a practical setting where the teacher specialties are totally disjoint, i.e., $Y_i \cap Y_j = \emptyset, \forall i \neq j$, as merging teachers with overlapping classes can be easily converted in to the disjoint situation.
\subsection{Model Uncertainty--Aware Knowledge Integration}
% We focus on the knowledge transfer based prediction supervision since it is directly related to the classification performance.
As there are no annotated data available due to the data privacy issue, we need to construct supervision for guiding the student.
% Different from previous explorations in computer vision resorting to the alignments on the internal states between the student and the teachers, we propose a principled framework which is more generalizable for heterogeneous models.
Given a golden label distribution $\mathcal{T}(x)$ for each instance $x$ over $Y$, we can train the student by minimizing the KL-divergence:
%  between its prediction and the estimated distribution
\begin{equation}
    \mathcal{L} = \sum_{x \in \mathcal{D}} \text{KL} \left(  S\left(x\right)  ||  \mathcal{T} \left(x\right)   \right),  
\label{eq:kd_loss}
\end{equation}
where $S(x)$ denotes the output distribution of the student for input $x$.
As we only operate on the output distribution level, thus this framework is generalizable for PLMs that potentially differ in the model architectures and training data distribution.
To estimate the golden supervision $\mathcal{T}(x)$, we first derive the correlation between $\mathcal{T}(x)$ and the prediction $T_i(x)$ of teacher model $T_i$.
Specifically, as teacher $T_i$ specializes in label set $Y_i$, it can only predict $T_i( y \mid x )$ for instance $x$ when $ y \in Y_i $.
Therefore, the correlation between $T_i( y  \mid x)$ and global probability $\mathcal{T}( y \mid x)$ over the full class set can be derived as: % 
\begin{align}
% &=\mathcal{T}\left(y  \mid x,  y \notin Y_{-i}\right) \\
T_{i}(y  \mid x ) 
&=\mathcal{T}\left(y  \mid x, y \in Y_{i}\right) \\
&=\frac{\mathcal{T}\left(y , y \in Y_{i} \mid x \right)}{\mathcal{T}\left( y \in Y_{i} \mid x \right)} . \label{eq:relation}
% \\
% &=\frac{\mathcal{T}(y = y^* \mid x )}{\sum_{k \in Y_{i}} \mathcal{T}(y=k \mid x )} 
\end{align}
The above derivation indicates that we can recover the golden probability distribution by (1) getting the teacher predictions, and (2) estimating the denominator, which means how likely the instance $x$ lies in the teacher $T_i$ specialty $Y_i$.
As instances associated with classes not in $Y_i$ can be treated as the out-of-distribution data for the teacher $T_i$, the teacher predictions would be more uncertain about these instances than that of in-distribution instances~\citep{hendrycks17baseline}.
We thus propose to approximate the denominator in an opposite direction, i.e., estimating how likely the instance is not belong to teacher $T_i$ via model uncertainty.
Followingly, we first explore different uncertainty estimations for recovering the golden supervision, and then introduce how we incorporates teacher predictions according to the estimated uncertainty scores.
% In the following section, we introduce our \textbf{M}odel \textbf{U}ncertainty--aware \textbf{K}nowledge \textbf{A}malgamation~(MUKA) framework, which consists of estimations of the golden supervision according to model uncertainty and an instance re-weighting mechanism for handling instances with confusing supervisions. 
% Figure~\ref{fig:uka_framework} gives an overview of our proposal. 

\subsubsection{Uncertainty Estimation} % 
\begin{figure}[t!]
    \centering
    \includegraphics[width=0.99\linewidth]{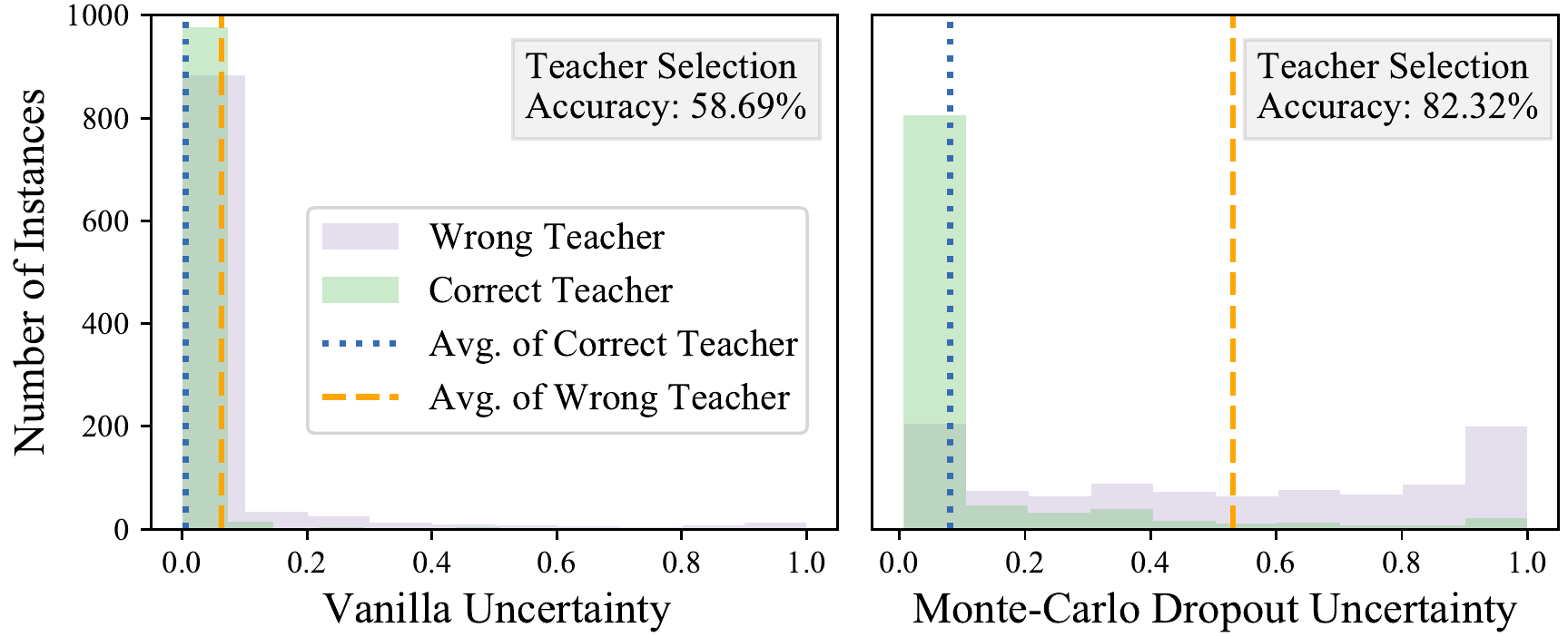}
    \caption{Model uncertainty~(normalized) distributions evaluated with $1000$ instances randomly sampled from the AG News dataset. The vanilla prediction entropy distributions of two teacher models overlaps greatly~(left), while Monte-Carlo Dropout produces a more accurate uncertainty approximation for distinguishing the correct teacher model~(right). Best viewed in color.
    }
    \label{fig:uncertainty_dist}
\end{figure}
% Instead of directly estimating this denominator, we approximate the probability that instance does not belong to any classes in $C_i$.
A na\"ive estimation is directly taking the statics like prediction entropy of predicted class distribution.
However, due to the over-confident issues of over-parameterized models like PLMs~\citep{guo2017calibration,desai2020calibration}, this simple estimation can be unreliable.
% Setup of this experiments 
We investigate this by first splitting the instances of the AG News dataset~\citep{cnn15zhang} into two sets with disjoint labels, and then fine-tuning teacher models on each set separately.
For each instance, there is a correct teacher that is capable of handling it and a wrong teacher that is not qualified for processing it.
We plot the prediction entropy distributions of the correct teacher and the wrong teacher in the left part of Figure~\ref{fig:uncertainty_dist}.
It can be found that the wrong teacher also produces confident predictions even for instances that are not in its speciality with nearly zero uncertainty scores, exhibiting a great overlap with the correct teacher model.
This indicates that utilizing the simple metric will mislead the identification of the adequate teacher.
To remedy this, inspired by recent progress in Bayesian neural networks~\citep{blundell2015weight,gal2016dropout}, we propose to add small perturbations to the model weights during inference to find out the correct teacher model.
The intuition behind is that, as the instance is well fitted by the parameter of the qualified teacher model, the teacher can produce confident results consistently in the multiple predictions even with small perturbed parameters. 
On the contrary, small perturbations on the model weights of the wrong teacher will lead to a drastic change in the output probabilities, resulting in more uncertain predictions on average.
Therefore, we can estimate the model uncertainty more accurately according to the average predictions under parameter perturbations. Specifically, we adopt Monte-Carlo Dropout~\citep{gal2016dropout}, where the output distribution of an instance $x$ with $T_i$ is calculated as:
\begin{align}
p_i\left(y \mid x, \mathcal{D}\right)&\approx \frac{1}{K}\sum_{k=1}^{K} p_i\left(y \mid \mathbf{W}_{k}^i, x\right) \\
& = \frac{1}{K} \sum_{k=1}^K  T_i\left(x, \mathbf{W}_k^i \right),
\label{eq:mc_teacher}
\end{align}
where $\mathbf{W}_k^i$ is the $k$-th masked weights of $T_i$ sampled from the Dropout distribution~\citep{srivastava2014dropout}, and $K$ is the sampling number.
The model uncertainty of teacher model $T_i$ thus can be summarized as the entropy of the averaged probability distribution $p_i$:
\begin{equation}
    u_i = H \left( p_i\right) = - \sum_{y=1}^{ |Y_i|} p_i^y \log p_i^y.
    \end{equation}
As shown in the right part of Figure~\ref{fig:uncertainty_dist}, the uncertainty distributions of the correct teacher and the wrong teacher model estimated via Monte-Carlo Dropout exhibit a clearer difference than vanilla prediction entropy, indicating its great potential for guiding the probability combination.

\begin{figure*}[t!]
    \centering
    \includegraphics[width=0.9\linewidth]{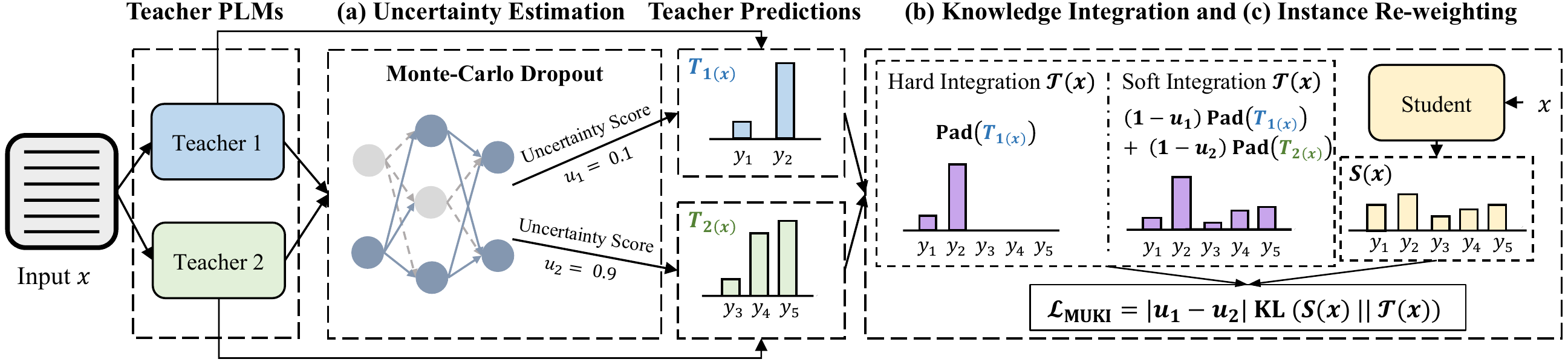}
    \caption{Overview of the proposed MUKI framework, which consists of (a) Model uncertainty scores estimation with Monte-Carlo Dropout. (b) Knowledge integration for estimating the golden supervision according to uncertainty scores with (c) instance-wise re-weighting mechanism.
    % The student model is trained via minimizing the KL-divergence between the hard(red) /soft(blue) estimated supervision and the student output, weighted by the model uncertainty margin of teachers. 
    Best viewed in color.
    }
    \label{fig:uka_framework}
\end{figure*}

\subsubsection{Knowledge Integration}
With the accurately estimated teacher uncertainties $U = \{u_1, \dots, u_N \}$ for each instance at hand, we design two methods for approximating the golden supervision to guide the student model:

\noindent\textbf{MUKI-Hard} which directly takes the supervision provided by the teacher with the lowest uncertainty as the golden distribution:
\begin{align}
\small 
i^* &= \mathop{\arg\min}_i u_i/\log | Y_i | \\ 
    \mathcal{T}(x) &\approx \text{Pad} \left( T_{i^*}\left(x\right)  \right) \\ 
    \text{Pad} \left( T_{i^*} \left(y \mid x\right) \right) &= \left\{
\begin{array}{cl}
T_{i^*}( y \mid x )  &  y \in Y_{i^*} \\
0  &  y \in Y - Y_{i^*}
\end{array} \right.
\end{align}
where $\log | Y_i | $ is a normalizing factor. 
%  for comparison between teacher models with a different number of classes
As $T_{i^*}$ only provides the label relation over the class set $Y_{i^*}$, the probabilities of classes not in $Y_{i^*}$ are set to zeros, denoted by the $\text{Pad}$ operation.
In this way, we are actually set $\mathcal{T}(y \in Y_{i^*} \mid x ) =1$, and thus the student can learn from the teacher model that is most confident about $x$.

\noindent\textbf{MUKI-Soft} which estimates the golden supervision as a weighted sum of teacher model predictions by taking the relative uncertain level into consideration:
\begin{align}
\small 
c_i &= 1 - u_i/\log |Y_i |  \\ 
w_i &= \frac{\exp\left(  c_i/ \tau\right)}{ \sum_{j=1}^N \exp(c_j/ \tau )} \label{eq:tau} \\
\mathcal{T}(x) &\approx \sum_{i=1}^{N}  w_i \text{Pad} \left(  T_i \left( x \right) \right) 
\end{align}
where $c_i$ denotes the confidence score which indicates how likely $x$ belongs to $C_i$, and $\tau$ is a hyper-parameter for controlling the smoothness of the weights.
In this way, the teacher with a higher confidence score contributes more to the estimated golden supervision signal.
Besides, the difference between the confidence scores reflects the inner correlation between the classes in different label groups, thus providing extra information for the classes in disjoint label sets. 

\subsubsection{Instance Re-weighting}
% Properly overlap between uncertainty scores 
% We notice that the uncertainty score
Furthermore, the uncertainty distribution overlapping in the right part of Figure~\ref{fig:uncertainty_dist} indicates that there is still a small portion of instances on which teacher models achieve similar confidence levels with Monte-Carlo Dropout.
For these instances, MUKI-Hard may wrongly select the supervision source, and MUKI-Soft would assign close weights to all teacher predictions, thus providing a vague even conflicting supervision signal.
% The estimated supervision of these instances can thus be misleading for the student model.
To remedy this, we devise an instance re-weighting mechanism by modifying the objective in Eq.~(\ref{eq:kd_loss}):
\begin{align}
    \mathcal{L}_{\text{MUKI}} &=\sum_{x \in \mathcal{D}} v\left(x\right) \text{KL} \left(  S\left(x\right) || \mathcal{T}(x)\right) \\ 
    v\left(x\right) &=  c_{\text{max}} - c_{\text{sec}} ,
\end{align}
where $c_{\text{max}}$ and $c_{\text{sec}}$ denotes the largest and the second large teacher confidence score for instance $x$, respectively.
By minimizing the instance-level weighted objective, the student is encouraged to focus more on the pivotal instances with clearer supervision signals, thus reducing the effect of potential confusing instances.
% achieving a better learning performance.

In summary, MUKI consists of supervision estimation and instance re-weighting mechanism based on model uncertainty for integrating the knowledge from different teachers.
Figure~\ref{fig:uka_framework} gives an overview of MUKI framework.
\section{Experiments}
% In this section, we first introduce the experimental setup including datasets used, compared methods and implementation details, followed by the main results and ablation studies.
% Finally, we explore MUKA in challenging settings, including merging multiple teachers models, heterogeneous teachers and cross-datasets teachers.
\begin{table*}[t!]
    \centering
    \small 
    \begin{tabular}{@{}l| c | c c  c  c |c@{}}
    \toprule
       \textbf{Method}  &  \textbf{Model Size} & \textbf{AG News}  & \textbf{THUCNews} & \textbf{Google Snippets}& \textbf{5Abstracts Group}  & \textbf{Average}\\
       \midrule 
           Supervised & 110M  & 94.6 $\pm$ 0.00&  97.8 $\pm$ 0.00 & 89.3 $\pm$ 0.00&  90.7 $\pm$ 0.00 &  93.10 \\ 
      \midrule
     Teacher 1  & 110M   & 49.9 $\pm$ 0.00 &48.8 $\pm$ 0.00 & 50.2 $\pm$ 0.00 &42.0  $\pm$ 0.00 & 47.73\\ 
      Teacher 2 & 110M  & 47.5 $\pm$ 0.00 &49.8 $\pm$ 0.00 & 43.5 $\pm$ 0.00& 51.5 $\pm$ 0.00 & 48.08\\
      Ensemble &  220M & 59.8 $\pm$ 0.00& 93.1 $\pm$ 0.00  &80.4 $\pm$ 0.00  &  62.3 $\pm$ 0.00 &  73.90\\
      \midrule 
       Vanilla KD &  110M  &63.1 $\pm$  0.81 & 94.9 $\pm$ 0.18 & 83.7  $\pm$ 1.30 & 67.0 $\pm$ 1.14 &  77.18 \\
       DFA &  110M & 66.4 $\pm$ 2.33 & 94.4 $\pm$ 0.22 & 82.6 $\pm$ 0.18 & 57.7 $\pm$ 4.41 & 72.78\\ 
       CFL&  110M  & 61.4  $\pm$  1.18 & 95.1  $\pm$ 0.21& 84.5  $\pm$ 0.45&61.6  $\pm$ 0.12&  75.65 \\ 
       UHC &  110M  & 78.8 $\pm$  1.42& 92.1 $\pm$ 0.63&86.3 $\pm$ 0.39& 71.4 $\pm$ 0.67 &  82.15\\  
       \midrule 
       MUKI-Hard~(Ours) &  110M  &  87.0  $\pm$ 0.40 &  \textbf{97.2}  $\pm$ 0.12 & \textbf{88.4}  $\pm$  0.32&  79.0  $\pm$ 0.82 & \textbf{87.90}\\ 
        MUKI-Soft~(Ours) &  110M  & \textbf{87.1}  $\pm$ 0.19&  \textbf{97.2}  $\pm$ 0.08 & 87.9 $\pm$ 0.32 &  \textbf{79.3} $\pm$  0.85 & 87.88\\ 
    \bottomrule
    \end{tabular}
    \caption{Comparisons on the benchmark datasets. 
    The results are classification accuracy averaged by three seeds, and standard deviations are reported. Both MUKI variants achieve statistically significant improvements over the best-performing baselines~($p < 0.01$). Best results are shown in bold. }
    \label{tab:main_ret}
\end{table*}
\subsection{Experimental Settings}
\paragraph{Datasets}
We conduct evaluations on four classic text classification benchmarks, including three English datasets: AG News~\citep{cnn15zhang}, Google Snippets~\citep{phan08gs}, 5Abstracts Group~\citep{liu-etal-2018-task}, and a Chinese dataset THUCNews~\citep{sun2016thuctc}. 
% For datasets without a validation set, we split 5\% of data from the training set to form a validation set for model selection.
We randomly split 5\% of data from the training set for datasets without a validation set to form a validation set for model selection.
The statistics of datasets can be found in Appendix~\ref{apx:dataset}.
\paragraph{Compared Methods}
We implement various baselines to evaluate our proposal, as follows:

\emph{Simple Baselines}, which require no additional training, including:
(1) Original Teacher: The teacher models are used independently for prediction. We set the probabilities of classes out of the teacher speciality to zeros.
(2) Ensemble: The output logits of teachers are directly concatenated for predictions over the union label set.

\emph{Distillation Methods}, which assume internal states of the teacher model are available and the student is trained via aligning the states of teacher models on  $\mathcal{D}$, including:
(1) Vanilla KD~\citep{Hinton2015Distilling}: The student is trained to mimic the soft targets produced by logits combination of all teacher models, via minimizing the vanilla KL-divergence objective. 
(2) DFA~\citep{shen19comprehensive}: DFA designs a layer-wise feature adaptation mechanism for providing extra guidance based on Vanilla KD. The student aligns its features to the merged features of multiple teachers layer by layer.
(3) CFL~\citep{Luo19CFL}: CFL first maps the hidden representations of the student and the teachers into a common space. The student is trained by aligning the mapped features to that of the teachers, with supplemental supervision from the logits combination. 
(4) UHC~\citep{Vongkulbhisal19UHC}: UHC splits the student logits into subsets corresponding to the class sets of teacher models. Each subset is trained to mimic the corresponding output of the teacher model.

A supervised learning method with labeled data is also included, to serve as a performance upper-bound for better understanding of the results.

\paragraph{Implementation Details}
We implement our framework using the HuggingFace transformers library~\citep{wolf-etal-2020-transformers}.
% We set the teacher number $N$ to $2$.
In our main setting, we set the teacher number $N$ to $2$ and we explore integrating multiple teachers in Section~\ref{subsec:chanllenge_setting}. 
For each dataset, the classes are randomly split into two non-overlapping parts, and two teachers are fine-tuned on each set separately to imitate the actual applications.
Detailed class split can be found in Appendix~\ref{apx:dataset}. 
The teacher and student models for English datasets and THUCNews are BERT-base-uncased~\citep{devlin2019bert} and BERT-wwm-ext~\citep{cui2020revisiting}, respectively.
% The student model is trained over the whole training data with label information removed and only the predictions from teacher models are used as supervision.
We first fine-tune the teacher models with the split labeled data for $3$ epochs with a learning rate $2\times10^{-5}$. The trained teacher model weights are frozen during the student training process.
We set the forward number $K$ of Monte-Carlo Dropout uncertainty estimation to $16$ and the dropout rate is set to $0.1$.
% , which is identical to the original BERT configuration
Temperature $\tau$ in Eq.~(\ref{eq:tau}) is set to $0.2$ according to our hyper-parameter analysis results in Appendix~\ref{apx:temp}.
The student model then is learned by optimizing the KL-divergence objective for $3$ epochs, with a $2\times10^{-5}$ learning rate and $32$ batch size.
The student is evaluated on the validation set every $100$ step. We select the best performing checkpoints for final evaluation.
The results are replicated with $3$ random seeds and we report the averaged accuracy.

\subsection{Main Results}
% \section{Analysis}
% In this section, we conduct experiments for validating the effectiveness of MUKA.
% In this section, we first present comparison results on the benchmark datasets by merging from two teachers, followed by explorations on the effects of MUKA components. Further, we investigate MUKA in challenging settings, including KA with multiple teachers, heterogeneous teachers with different architectures, and cross-dataset teachers.

% We first present comparison results on the benchmark dataset, then explore the effect of UKA components, and further investigate UKA in more challenging settings like multiple teacher models and teacher models with heterogeneous architectures.

% \subsubsection{Main Results}
The model performance comparison on the four datasets and the corresponding model size are listed in Table~\ref{tab:main_ret}.
% We observe that methods based on supervision on the combinated logits 
Our findings are: 
(1) Simple baselines fall far behind, showing that it is necessary to design a proper integration strategy for amalgamating the knowledge from different teachers. 
(2) While extra feature alignment objectives are adopted, DFA and CFL cannot achieve consistent improvements over Vanilla KD. We speculate the reason is that the supervision based on feature alignments is unstable, as teacher features are fine-tuned for specializing in different semantic classes.
(3) UHC achieves better average results than Vanilla KD, while performing relatively worse on the THUCNews dataset. It indicates there exists potential supervision conflict as UHC matches the student output independently to that of teachers, thus limiting its generalizability on different datasets.
(4) Two variants of MUKI both significantly outperform the previous baseline models on all the datasets, and the average accuracy of MUKI-Hard is achieves a $5.75$ points gain over the best performing baseline model. On the THUCNews dataset, while no label information is included during the knowledge amalgamation, MUKI can obtain a $97.2$ accuracy, which is very close to $97.8$ of the supervised learning method.
We attribute the success to that MUKI provides the student with the accurately estimated golden probability distribution over the union label set according to model uncertainty, which can effectively transfer the knowledge and alleviate potential supervision conflicting.
These promising results indicate that our MUKI framework can produce better supervision for training the student model, thus has great potentials for reusing PLMs with different label sets.

\subsection{Ablation Studies}
\begin{table}[t!]
    \centering
    \small 
    \begin{tabular}{@{}lc c@{}}
    \toprule 
    \textbf{Method}     &  \textbf{AG News}& \textbf{THUCNews} \\
    \midrule
    MUKI-Hard & 87.0 $\pm$  0.40 &  97.2 $\pm$ 0.12 \\ 
    \quad w/o Monte-Carlo Dropout& 65.1 $\pm$ 1.67 &97.0 $\pm$ 0.13\\ 
    \quad w/o Instance Re-weighting& 85.5  $\pm$ 0.51 &  96.9 $\pm$ 0.06  \\ 
    \midrule 
    MUKI-Soft & 87.1  $\pm$ 0.19 &  97.2  $\pm$  0.08 \\ 
    \quad w/o Monte-Carlo Dropout&  74.2  $\pm$ 0.28 &95.4  $\pm$ 0.20 \\ 
    \quad w/o Instance Re-weighting  &  86.7  $\pm$ 0.30 &  96.8  $\pm$ 0.03 \\ 
    \bottomrule
    \end{tabular}
    \caption{Ablation analysis of MUKI. The removed modules both lead to deteriorated performance.}
    % , validating the effectiveness of accurate model uncertainty estimation and instance re-weighting mechanism.}
    \label{tab:ablation_study}
\end{table}

% \paragraph{Contribution of Components in MUKI} 
We conduct ablation experiments on two large datasets, i.e., AG News and THUCNews for stable results, to explore the following two questions.

\noindent\textbf{How Monte-Carlo Dropout benefits knowledge source identification?}
We replace the Monte-Carlo Dropout estimation of model estimation with a single forward estimation, i.e., setting $K=1$ in Eq.~(\ref{eq:mc_teacher}).
As shown in Table~\ref{tab:ablation_study}, we find that the performance is degraded on both datasets.
% This result validates our motivation to adopt the Monte-Carlo Dropout method to better estimate the model uncertainty.
% 为什么在 AG News 上效果好
Interestingly, we find that the accuracy drop is much clearer on the AG News than that on the THUCNews.
To explore this, we compute the average ECE score~\citep{guo2017calibration} of two teacher models on the out-of-distribution samples, where higher ECE scores indicate more severe over-confident predictions.
The teacher models of AG News achieve an average $45.42$ ECE score, while that of THUCNews is $19.52$.
Therefore, the teacher models of AG News exhibit a much more serious over-confident issue than that of THUCNews.
This result verifies that our adoption of the Monte-Carlo Dropout technique is effective for accurately identifying the adequate teacher model, especially when teachers tend to make over-confident predictions.

\begin{figure}[t]
    \centering
    \includegraphics[width=0.75\linewidth]{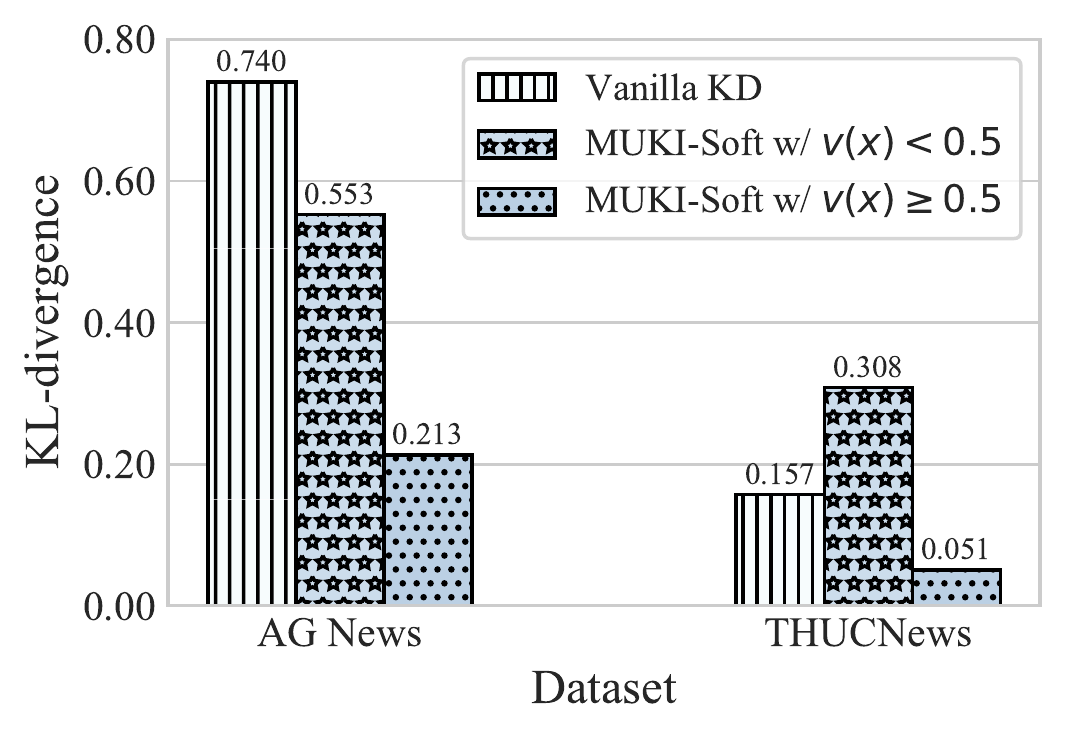}
    \caption{Supervision quality measured by the KL-divergence to the golden supervision of different methods. For MUKI-Soft, we divide instances into two groups according to the uncertainty margin $v(x)$. }
    \label{fig:kl_analysis}
\end{figure}

\noindent\textbf{How instance-wise re-weighting benefits supervision integration?} 
In Table~\ref{tab:ablation_study}, we find that removing the instance re-weighting mechanism leads to deteriorated results for both MUKI variants.
% , demonstrating that paying more attention to instances with clearer supervision is beneficial for the student
We further probe whether the re-weighting mechanism is capable of resolving the vague supervision issue when teachers achieve similar uncertainty scores. 
Specifically, we train a PLM with all labeled data as the proxy of an oracle model, which thus can provide golden supervision over the union label set.
We then calculate the KL-divergence between the golden probability distribution and the approximated one of different combination methods. 
Lower KL-divergence indicates the combined predictions are more correct.
We discard results of MUKI-Hard as KL-divergence is not defined for distributions with zeros, and divide MUKI-Soft into two groups according to the uncertainty score margin $v(x)$, i.e. instances with $v(x) \geq 0.5$ and that with $v(x) < 0.5 $.

As shown in Figure~\ref{fig:kl_analysis},
% The bottom two rows are results of two parts of the instances,.
we find that the predictions of instances with $v(x) \geq 0.5$, are much closer to the golden distributions than the Vanilla KD.
This indicates that the estimated supervision of instances with a clearer confidence margin is of higher quality, thus paying more attention to these instances is effective for knowledge integration.

\begin{table}[t!]
    \centering
    \small 
    \begin{tabular}{@{}lc cc@{}}
    \toprule 
    \multirow{2}{*}{\textbf{Method}}     &  \textbf{3 Teachers}& \textbf{4 Teachers} & \textbf{5 Teachers}  \\
     & \{3,3,4\} & \{2,2,2,4\} & \{2,2,2,2,2\}\\ 
    \midrule
    Teacher 1 & 29.9 $\pm$ 0.00  & 20.0 $\pm$ 0.00 & 20.0 $\pm$ 0.00 \\
    Teacher 2 & 29.8 $\pm$ 0.00&  19.1 $\pm$ 0.00 & 19.1 $\pm$ 0.00\\
    Teacher 3 & 39.8 $\pm$ 0.00& 19.9 $\pm$ 0.00& 19.9 $\pm$ 0.00\\
    Teacher 4 & N / A & 39.8 $\pm$ 0.00& 20.0 $\pm$ 0.00\\
    Teacher 5& N / A & N / A & 20.0 $\pm$ 0.00\\
    Ensemble &  91.0 $\pm$ 0.00 & 74.0 $\pm$ 0.00& 80.6 $\pm$ 0.00\\
    \midrule 
    Vanilla KD  &92.5 $\pm$ 1.13 &76.5 $\pm$  0.57& 82.6 $\pm$ 1.64\\ 
     DFA &  91.1 $\pm$	1.25& 79.6 $\pm$	1.34& 82.0 $\pm$		2.04\\ 
    CFL &  93.9 $\pm$ 1.07&77.4  $\pm$ 1.34& 84.0 $\pm$ 0.24\\
    UHC & 84.4 $\pm$ 0.91& 71.6  $\pm$ 2.86 & 69.4 $\pm$ 2.73\\ 
    \midrule 
    MUKI-Hard & \textbf{94.7} $\pm$ 0.20 &93.4$^*$\hspace{-\lsuperstar}  $\pm$ 0.58& \textbf{90.5}$^*$\hspace{-\lsuperstar}  $\pm$ 0.32\\ 
    MUKI-Soft & \textbf{94.7} $\pm$ 0.14 & \textbf{93.6}$^*$\hspace{-\lsuperstar}  $\pm$ 0.30& \textbf{90.5}$^*$\hspace{-\lsuperstar} $\pm$ 1.13 \\ 
    \bottomrule
    \end{tabular}
    \caption{Results of merging multiple teacher models on the THUCNews dataset. $^*$ denotes the improvement over the best performing baseline is significant~($p < 0.05$). N / A means that the teacher model does not exist in the corresponding setting.
    % The best results are shown in bold and
    }
    \label{tab:multiple_teacher}
\end{table}

\subsection{Results in Challenging Settings}

\label{subsec:chanllenge_setting}
\paragraph{KI with Multiple Teachers} As MUKI is agnostic to the number of teacher models, we explore its adaptability with more teacher models.
We conduct experiments on the THUCNews dataset as it has $10$ classes, allowing us to train up to $5$ teacher models specialized in different class sets.
The $10$ classes are split into $\{3, 3, 4\}$, $\{2, 2, 2, 4\}$ and $\{2, 2, 2, 2, 2\}$ for $3$, $4$ and $5$ teachers, respectively.
As shown in Table~\ref{tab:multiple_teacher}, the proposed MUKI framework generalizes well to this setting, outperforming previous baselines with a clear margin.
% the model performance is decreased as the number of teacher models increases. 
% However, different from UHC which performs well under the 2-teacher scenario while suffers when merging from many teacher models, our UKA generalizes well with little performance degradation with increased number of teacher models.
% 写一下当 label gap 大的时候差距很大
Besides, we find that all baselines based on logits alignment perform poorly under the $4$-teacher scenario.
We attribute it to that when some teacher models have more classes than others, they usually produce a larger range of logits to make the prediction more distinguishable.
Directly combing the teacher logits thus leads to a biased probability distribution. 
Our MUKI instead estimates the golden supervision according to model uncertainty scores, thus producing better supervision even teacher models exhibit different logit scales.

\paragraph{KI with Heterogeneous Teachers}
As MUKI only operates on the output distribution level, it is generalizable for heterogeneous teachers. We verify this by merging teachers with different model structures.
Specifically, we adopt BERT-base~($12$ layers and $768$ hidden units) and BERT-large~($24$ layers and $1024$ hidden units) as the teachers, respectively. 
As shown in Table~\ref{tab:heter_arch}, we find that while a larger teacher tends to perform better, the student model performs worse on the THUCNews dataset than learning from two BERT-base teachers, indicating it is challenging to integrate knowledge in this setting.
Our MUKI achieves the best results on these two datasets, showing its effectiveness for heterogeneous teachers.
% It can be found that methods-based supervision concatenating the logits suffers when merging from teacher model with different architectures.
\begin{table}[t!]
    \centering
    \small 
    \begin{tabular}{@{}l c  c@{}}
    \toprule 
    \multirow{2}{*}{\textbf{Method}} &  \textbf{\emph{$\text{T}_1$: BERT-base}}   & \textbf{\emph{$\text{T}_2$: BERT-large}}    \\
    \cmidrule(lr){2-3}
    &  \textbf{AG News} & \textbf{THUCNews} \\ 
    \midrule
    Teacher 1 &49.7 $\pm$ 0.00  & 48.8 $\pm$ 0.00 \\
    Teacher 2 &  47.2 $\pm$ 0.00& 49.8 $\pm$ 0.00  \\
    Ensemble & 76.6 $\pm$ 0.00 & 79.3 $\pm$ 0.00\\
    \midrule 
    Vanilla KD  &79.6  $\pm$ 0.22 &82.9 $\pm$ 1.36\\ 
    DFA & 78.2 $\pm$	0.30 & 84.7 $\pm$ 1.74  \\ 
    CFL &  75.9 $\pm$ 0.63&  81.9 $ \pm$ 1.37 \\
    UHC & 78.3 $\pm$ 	2.65& 92.3  $\pm$  1.08\\ 
    \midrule 
    MUKI-Hard &78.3 $\pm$ 	1.58 & \textbf{95.4}$^*$\hspace{-\lsuperstar} $\pm$ 0.45\\ 
    MUKI-Soft & \textbf{80.6} $\pm$  0.17& \textbf{95.4}$^*$\hspace{-\lsuperstar} $\pm$  0.29\\ 
    \bottomrule
    \end{tabular}
    \caption{Results of merging BERT-base and BERT-large. $^*$ denotes results are statistically significant~($p < 0.05$).}
    \label{tab:heter_arch}
\end{table}
\paragraph{KI with Cross-Dataset Teachers}
% Besides, we conduct knowledge amalgamation for teacher models specialize in different source datasets.
Specifically, we fine-tune teacher models on different datasets separately and then train a student to perform classification over the union label set of both datasets.
The multilingual BERT-base is adopted for the teachers and the student in the cross-lingual setting.
The results of merging knowledge from two English datasets, AG News and Google-Snippets and even cross-lingual datasets, AG News~(in English) and THUCNews~(in Chinese) are listed in Table~\ref{tab:heter_data}.
We find that MUKI still outperforms previous baseline models in both settings.
Interestingly, we find that the MUKI-Hard is consistently better than MUKI-Soft in this setting. We speculate the reason is that the correlations between classes of different datasets are weak, thus modeling the label relation in these disjoint groups is unnecessary.
% and may even hurt the KI performance.

\begin{table}[t!]
    \centering
    \small 
    \begin{tabular}{@{}l c | c  @{}}
    \toprule 
    \multirow{2}{*}{\textbf{Method}} &  \textbf{\emph{$\text{T}_1$: AG News~(en)}}   & \textbf{\emph{$\text{T}_1$: AG News~(en)}} \\
    &  \textbf{\emph{$\text{T}_2$: GS~(en)}} &  \textbf{\emph{$\text{T}_2$: THUCNews~(zh)}} \\ 
    \midrule
    Teacher 1 &26.8 $\pm$ 0.00 &  0.50 $\pm$ 0.00\\
    Teacher 2 &  63.0 $\pm$ 0.00& 54.6 $\pm$ 0.00  \\
    Ensemble & 74.3 $\pm$ 0.00& 54.5 $\pm$ 0.00 \\
    \midrule 
    Vanilla KD  &75.1  $\pm$ 0.58 &   54.6  $\pm$ 0.05\\ 
    DFA & 74.7 $\pm$	0.30  & 54.1 $\pm$ 0.20 \\ 
    CFL & 74.0 $\pm$ 0.50 & 54.3  $\pm$ 0.65\\
    UHC & 72.0 $\pm$ 0.73& 53.5  $\pm$ 0.17\\ 
    \midrule 
    MUKI-Hard & \textbf{76.3} $\pm$ 0.54&  \textbf{68.1}$^*$\hspace{-\lsuperstar}  $\pm$ 0.20\\ 
    MUKI-Soft & 75.9 $\pm$  0.47& 65.6$^*$\hspace{-\lsuperstar}   $\pm$ 0.40\\ 
    \bottomrule
    \end{tabular}
    \caption{Cross-dataset results. GS is short for Google Snippets. 
    (Left Column) Integrating teachers major in AG News and GS, respectively. (Right Column) Merging teachers major in AG News~(in English) and THUCNews~(in Chinese), respectively. $^*$ denotes results are statistically significant with $p < 0.05$. }
    \label{tab:heter_data}
\end{table}

\subsection{Results for Structured Prediction}
\begin{table}[t!]
    \centering
    \small 
    \begin{tabular}{@{}l c | c  @{}}
    \toprule 
    \textbf{Method} &  \textbf{CoNLL 2003}   & \textbf{OntoNotes 5.0} \\
    \midrule 
    Teacher 1 & 55.4 $\pm$ 0.00 & 46.20 $\pm$ 0.00\\
    Teacher 2 & 38.6 $\pm$ 0.00&  25.34 $\pm$ 0.00  \\
    Ensemble & 79.5 $\pm$ 0.00& 49.85 $\pm$ 0.00 \\
    \midrule 
    Vanilla KD  & 83.6 $\pm$ 0.66  &  53.33 $\pm$ 0.00\\
    DFA &  83.9 $\pm$ 0.45  &  52.66 $\pm$ 0.13 \\ 
    CFL & 83.1 $\pm$ 0.79 &   53.39 $\pm$  0.63\\
    UHC & 81.4 $\pm$ 0.24&  55.17 $\pm$ 	1.20\\ 
    \midrule 
    MUKI-Hard & \textbf{85.3}$^{*}$\hspace{-\lsuperstar} $\pm$ 0.26&  \textbf{59.83}$^{**}$\hspace{-\lsuperstar}  $\pm$ 0.31\\ 
    MUKI-Soft & 84.5 $\pm$  0.30&   59.33 $\pm$ 0.80 \\ 
    \bottomrule
    \end{tabular}
    \caption{F1-score results on two NER datasets. The results are statistically significant~($^*$ for $p <0.05$ and $^{**}$ for $p <0.01$). MUKI outperforms all the baselines significantly, verifying its generalizability for complicated tasks like structured prediction.}
    \label{tab:ner}
\end{table}

%%%	% %	%
%
%
%
%

We extend the KI framework into a classic structured prediction task, i.e., named entity recognition~(NER).
The problem is modeled as a tagging problem following \citet{devlin2019bert}, and we conduct evaluations on CoNLL 2003~\citep{sang2003conll} and OntoNotes 5.0~\citep{pradhan2013ontonotes}.
Specifically, we split the entity types of the dataset into two groups and train two teachers responsible for identifying the entities in each group, respectively. We refer readers to Appendix~\ref{apx:dataset} for  the detailed dataset statistics and the division of entity types.
% The KI is achieved by training a student model with unlabeled data for predicting all types of entities.
% Our framework can easily adapted to the structure predictions task by computing the model uncertainty at token-level.
We adapt MUKI to the NER task by estimating the model uncertainty and integrating the predictions at the token level.
Besides, we notice that the teacher will predict the non-entity tag with high confidence for tokens of entities it cannot handle.
% the O-tag in the IOB2 schema
Therefore, we adjust the uncertainty estimation procedure by calculating the entropy of probability distribution over the tags of entity types. 
The knowledge integration remains the same with the classification problem.
As shown in the in Table~\ref{tab:ner}, our MUKI still performs the best among all methods, validating that the proposed framework is generalizable for structured predictions tasks like NER. 
% The overall performance on OntoNotes is , which is consistent with previous findings that .

\section{Related Work}

Our work is mainly related to knowledge distillation~(KD), which aims to transfer the knowledge from a teacher model to a student model.
\citet{Hinton2015Distilling} utilize the soft labels of the teacher model for the student to learn,
and \citet{romero15fitnet} align the internal representations between the student and the teacher.
Recent studies apply KD for PLMs successfully by matching the intermediate states~\citep{Sun2019PatientKD,wang2020minilm} and enriching the training data with data augmentation~\citep{Jiao2019TinyBERT,liang2020mixkd}, learning from multiple teachers~\citep{wu-etal-2021-one}, and dynamically adjusting the learning objectives~\citep{li-etal-2021-dynamic}.
%  with data augmentation ,li-etal-2021-dynamic ,ren-etal-2021-text
Nevertheless, all these KD studies assume that the student has an identical label set with the target teacher model(s). Instead, knowledge integration removes this restriction by merging knowledge from multiple teacher with various label sets to train a versatile student model.

% of the teacher models specialized in different classification problems.
% \paragraph{Knowledge Amalgamation~(KA)} aims to train a versatile student model with unlabeled data, by amalgamating the knowledge from multiple teachers. % ~\citep{shen19comprehensive}
Recently, the idea of integrating knowledge from models with different skills has been explored in computer vision~\citep{shen19comprehensive,Ye19Master,Luo19CFL,Vongkulbhisal19UHC} and graph neural networks~\citep{jing2021gnnKA}, or extended to a semi-supervised setting~\citep{Thadajarassiri21SKA}.
% and can be roughly categorized into two streams.
% One is matching the student outputs separately to the corresponding teacher~\citep{Vongkulbhisal19UHC}.
% The other instead focuses on aligning the representations of the student and the teachers:
% \citet{shen19comprehensive} map the student features to that of the teacher model layer-by-layer and \citet{Luo19CFL} conduct the feature alignment in a common space by minimizing the maximum mean discrepancy.
% % However, these methods are sub-optimal due to the inherent over-confident problem and the possible heterogeneity of PLMs.
% However, the former results in conflicting supervision for the student as teacher-PLMs tend to be overconfident~\citep{desai2020calibration}, and the latter paradigm requires complicated optimization design for generalizing to heterogeneous teachers.
% KA has recently been extended to graph neural networks~\citep{jing2021gnnKA} and explored in a semi-supervised setting~\citep{Thadajarassiri21SKA}.
To the best of our knowledge, we are the first to explore knowledge integration for PLMs, which is of great practical value as there are abundant released PLMs.
Besides, different from previous methods relying on supervision from feature alignments~\citep{shen19comprehensive,Luo19CFL} or independent logits matching~\citep{Vongkulbhisal19UHC}, our MUKI framework operates on the distribution level by utilizing model uncertainty to approximate the golden supervision. Therefore, MUKI is more effective and generalizable for integrating knowledge from heterogeneous PLMs.
\section{Error Analysis}

% 
% My key concern is that the performance of MUKI (and all baselines) is worse than naive supervised learning models. Only using the supervision signals provided by teacher models to train student models does not make sense since it hurts the performance. 

% Could this be extended to teacher models that target scenarios more complex than classification, say for example span prediction or structured prediction? In what scenarios would MUKI fail or perform poorly, and what can be done by way of extension to overcome some of these problems? What happens when some teacher models are worse than others - can anything be done to mitigate the effects of weaker teachers? Beyond bias in PLM, are there any ethical concerns around leakage of private or proprietary information that need to be addressed in future work? This is not necessarily specific to MUKI but to KD in general, but the fact that MUKI deals with multiple teachers could be an aggravating factor.

\begin{figure}[t!]
    \centering
    \includegraphics[width=1.0\linewidth]{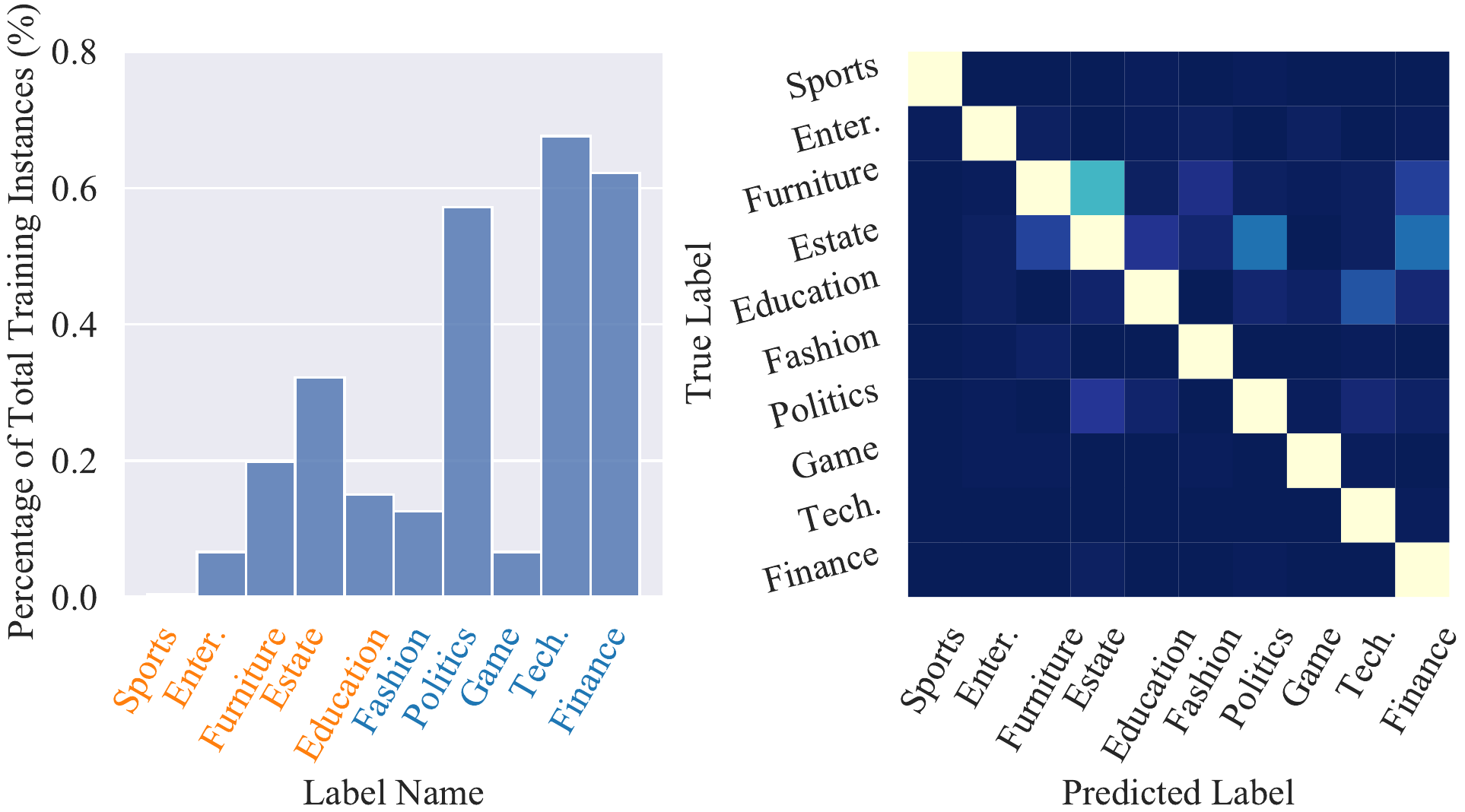}
    \caption{(Left) Label distribution of instances with wrongly selected teachers. Labels in the same color indicate they are in the same teacher specialty.
    (Right) The confusion matrix of an oracle model.}
    \label{fig:error_analysis}
\end{figure}

The MUKI is built on the assumption that the model uncertainty estimation can faithfully reflect the ability of teacher model.
We perform an error analysis to investigate when this assumption will fail, i.e, the estimated model uncertainty misguides the teacher selection.
% for identifying the adequate teacher model, we are curious in  which scenarios the teacher.
% We investigate the limitations of MUKI will selects a wrong teacher model.
Specifically, we probe the label distribution of instances on which MUKI assigns a higher uncertainty score to the correct teacher on THUCNews. 
As shown in the left part of Figure~\ref{fig:error_analysis}, the uncertainty-based teacher selection only fails on a small portion of training examples, i.e., $2.8$\% of total training instances. 
Interestingly, the labels of these instances are not uniformly distributed, e.g., \emph{Estate} and \emph{Tech.} have higher error rates than other classes.
We further plot the label confusion matrix of an oracle model that is fine-tuned with labeled data with all categories, in the right part of Figure~\ref{fig:error_analysis}.
We find that there are classes that can even confuse the oracle model, e.g., instances of \emph{Estate} are tending to be classified into \emph{Politics} and \emph{Finance}, indicating that the mis-identification is partially due to the inherent class similarity.
% On the other hand, while the oracle hardly makes wrong predictions on instances belonging to \emph{Tech.} and \emph{Finance}, the uncertainty estimation fails more frequently on these two classes than other classes.
These findings suggest that the performance MUKI can be limited when the integrating teacher models whose classes are highly correlated.
% We leave the investigation on this phenomenon for future work.
As a remedy, the proposed instance re-weighting mechanism is an effective ad-hoc strategy, as the average teacher uncertainty margin $v(x)$ is $0.04$, which can greatly reduce the negative impact of these instances. Developing better model uncertainty estimation techniques like incorporating more class information into the estimation process is also promising.
Besides, the uncertainty estimation can also be influenced by the model capacity. As deeper models tend to produce more confident predictions, the model selection based on uncertainty scores will favor stronger teacher models when there exist particularly weak teacher models. In such a case, applying model calibration techniques like temperature scaling according to the teacher model size~\citep{guo2017calibration,desai2020calibration} before estimating the uncertainty can be beneficial.
% Developing better model uncertainty estimation techniques like incorporating more class information while in the estimation process is also promising for the potential issue.
% Besides, developing better techniques for more accurate model uncertainty can also be explored to reduce the.

\section{Conclusion}
In this paper, we explore knowledge integration for PLMs to promote better model reuse.
% We estimate the probability distribution over the union label set for .
We present MUKI, which integrates teacher predictions according to the model uncertainty estimated via Monte-Carlo Dropout, and dynamically adjusts the instance contribution according to the uncertainty margin.
Extensive results on benchmark datasets demonstrate that MUKI can substantially outperform strong baselines, and perform well in challenging settings such as merging heterogeneous teachers.
Further investigation shows that MUKI can be extended to sequence labeling. In the future, we are interested in developing better integration frameworks for more complex tasks.
\section{Ethical Considerations}
Our work faces several ethical challenges.
% r, finetuned LMs may exhibit biases against certain protected groups
As the released PLMs may exhibit potential biases against specific groups, e.g., gender or ethnic minorities~\citep{kurita2019measuring,kennedy2020contextualizing}, these social biases can be propagated to the merged student model. 
Besides, users may collect unlabeled data from the web for conducting knowledge integration, which possibly contains offensive content and thus introduces new biases into the merged student model as well.
% Data privacy 的 issue 
We offer possible remedies to reduce the concerns.
For the biases exhibited in the teacher PLMs, de-biasing techniques~\citep{zmigrod2019cda,liang2020towards,schick2020self} can be applied to eliminate the potential biases in the teachers before  integration.
For the offensive unlabeled data collected from the internet, simple template-based or human-in-the-loop data cleaning strategies can be adopted, to identify and filter potential biased data.
Except for these techniques, developing a bias-aware knowledge transfer framework that can de-bias the supervision for the student model while maintaining task performance is also promising~\citep{Gupta2022MitigatingGB}.

\section*{Acknowledgements}
We thank all the anonymous reviewers for their constructive comments, and Shuhuai Ren for his valuable suggestions in preparing the manuscript.
This work was supported by a Tencent Research Grant. 
Xu Sun is the corresponding author.

\bibliography{anthology,custom}

\begin{thebibliography}{39}
\expandafter\ifx\csname natexlab\endcsname\relax\def\natexlab#1{#1}\fi

\bibitem[{Blundell et~al.(2015)Blundell, Cornebise, Kavukcuoglu, and
  Wierstra}]{blundell2015weight}
Charles Blundell, Julien Cornebise, Koray Kavukcuoglu, and Daan Wierstra. 2015.
\newblock \href {http://proceedings.mlr.press/v37/blundell15.html} {Weight
  uncertainty in neural network}.
\newblock In \emph{Proceedings of the 32nd International Conference on Machine
  Learning, {ICML} 2015, Lille, France, 6-11 July 2015}, volume~37 of
  \emph{{JMLR} Workshop and Conference Proceedings}, pages 1613--1622.
  JMLR.org.

\bibitem[{Cui et~al.(2020)Cui, Che, Liu, Qin, Wang, and Hu}]{cui2020revisiting}
Yiming Cui, Wanxiang Che, Ting Liu, Bing Qin, Shijin Wang, and Guoping Hu.
  2020.
\newblock \href {https://doi.org/10.18653/v1/2020.findings-emnlp.58}
  {Revisiting pre-trained models for {C}hinese natural language processing}.
\newblock In \emph{Findings of the Association for Computational Linguistics:
  EMNLP 2020}, pages 657--668, Online. Association for Computational
  Linguistics.

\bibitem[{Desai and Durrett(2020)}]{desai2020calibration}
Shrey Desai and Greg Durrett. 2020.
\newblock \href {https://doi.org/10.18653/v1/2020.emnlp-main.21} {Calibration
  of pre-trained transformers}.
\newblock In \emph{Proceedings of the 2020 Conference on Empirical Methods in
  Natural Language Processing (EMNLP)}, pages 295--302, Online. Association for
  Computational Linguistics.

\bibitem[{Devlin et~al.(2019)Devlin, Chang, Lee, and
  Toutanova}]{devlin2019bert}
Jacob Devlin, Ming-Wei Chang, Kenton Lee, and Kristina Toutanova. 2019.
\newblock \href {https://doi.org/10.18653/v1/N19-1423} {{BERT}: Pre-training of
  deep bidirectional transformers for language understanding}.
\newblock In \emph{Proceedings of the 2019 Conference of the North {A}merican
  Chapter of the Association for Computational Linguistics: Human Language
  Technologies, Volume 1 (Long and Short Papers)}, pages 4171--4186,
  Minneapolis, Minnesota. Association for Computational Linguistics.

\bibitem[{Gal and Ghahramani(2016)}]{gal2016dropout}
Yarin Gal and Zoubin Ghahramani. 2016.
\newblock \href {http://proceedings.mlr.press/v48/gal16.html} {Dropout as a
  bayesian approximation: Representing model uncertainty in deep learning}.
\newblock In \emph{Proceedings of the 33nd International Conference on Machine
  Learning, {ICML} 2016, New York City, NY, USA, June 19-24, 2016}, volume~48
  of \emph{{JMLR} Workshop and Conference Proceedings}, pages 1050--1059.
  JMLR.org.

\bibitem[{Guo et~al.(2017)Guo, Pleiss, Sun, and
  Weinberger}]{guo2017calibration}
Chuan Guo, Geoff Pleiss, Yu~Sun, and Kilian~Q. Weinberger. 2017.
\newblock \href {http://proceedings.mlr.press/v70/guo17a.html} {On calibration
  of modern neural networks}.
\newblock In \emph{Proceedings of the 34th International Conference on Machine
  Learning, {ICML} 2017, Sydney, NSW, Australia, 6-11 August 2017}, volume~70
  of \emph{Proceedings of Machine Learning Research}, pages 1321--1330. {PMLR}.

\bibitem[{Gupta et~al.(2022)Gupta, Dhamala, Kumar, Verma, Pruksachatkun,
  Krishna, Gupta, Chang, Steeg, and Galstyan}]{Gupta2022MitigatingGB}
Umang Gupta, J.~Dhamala, Varun Kumar, Apurv Verma, Yada Pruksachatkun,
  Satyapriya Krishna, Rahul Gupta, Kai-Wei Chang, Greg~Ver Steeg, and A.~G.
  Galstyan. 2022.
\newblock Mitigating gender bias in distilled language models via
  counterfactual role reversal.
\newblock \emph{ArXiv}, abs/2203.12574.

\bibitem[{Hendrycks and Gimpel(2017)}]{hendrycks17baseline}
Dan Hendrycks and Kevin Gimpel. 2017.
\newblock \href {https://openreview.net/forum?id=Hkg4TI9xl} {A baseline for
  detecting misclassified and out-of-distribution examples in neural networks}.
\newblock In \emph{5th International Conference on Learning Representations,
  {ICLR} 2017, Toulon, France, April 24-26, 2017, Conference Track
  Proceedings}. OpenReview.net.

\bibitem[{Hinton et~al.(2015)Hinton, Vinyals, and Dean}]{Hinton2015Distilling}
Geoffrey Hinton, Oriol Vinyals, and Jeff Dean. 2015.
\newblock \href {https://arxiv.org/abs/1503.02531} {Distilling the knowledge in
  a neural network}.
\newblock \emph{ArXiv preprint}, abs/1503.02531.

\bibitem[{Jiao et~al.(2020)Jiao, Yin, Shang, Jiang, Chen, Li, Wang, and
  Liu}]{Jiao2019TinyBERT}
Xiaoqi Jiao, Yichun Yin, Lifeng Shang, Xin Jiang, Xiao Chen, Linlin Li, Fang
  Wang, and Qun Liu. 2020.
\newblock \href {https://doi.org/10.18653/v1/2020.findings-emnlp.372}
  {{T}iny{BERT}: Distilling {BERT} for natural language understanding}.
\newblock In \emph{Findings of the Association for Computational Linguistics:
  EMNLP 2020}, pages 4163--4174, Online. Association for Computational
  Linguistics.

\bibitem[{Jing et~al.(2021)Jing, Yang, Wang, Song, and Tao}]{jing2021gnnKA}
Yongcheng Jing, Yiding Yang, Xinchao Wang, Mingli Song, and Dacheng Tao. 2021.
\newblock Amalgamating knowledge from heterogeneous graph neural networks.
\newblock In \emph{CVPR}, pages 15709--15718.

\bibitem[{Kennedy et~al.(2020)Kennedy, Jin, Mostafazadeh~Davani, Dehghani, and
  Ren}]{kennedy2020contextualizing}
Brendan Kennedy, Xisen Jin, Aida Mostafazadeh~Davani, Morteza Dehghani, and
  Xiang Ren. 2020.
\newblock \href {https://doi.org/10.18653/v1/2020.acl-main.483}
  {Contextualizing hate speech classifiers with post-hoc explanation}.
\newblock In \emph{Proceedings of the 58th Annual Meeting of the Association
  for Computational Linguistics}, pages 5435--5442, Online. Association for
  Computational Linguistics.

\bibitem[{Kurita et~al.(2019)Kurita, Vyas, Pareek, Black, and
  Tsvetkov}]{kurita2019measuring}
Keita Kurita, Nidhi Vyas, Ayush Pareek, Alan~W Black, and Yulia Tsvetkov. 2019.
\newblock \href {https://doi.org/10.18653/v1/W19-3823} {Measuring bias in
  contextualized word representations}.
\newblock In \emph{Proceedings of the First Workshop on Gender Bias in Natural
  Language Processing}, pages 166--172, Florence, Italy. Association for
  Computational Linguistics.

\bibitem[{Li et~al.(2021)Li, Lin, Ren, Li, Zhou, and
  Sun}]{li-etal-2021-dynamic}
Lei Li, Yankai Lin, Shuhuai Ren, Peng Li, Jie Zhou, and Xu~Sun. 2021.
\newblock \href {https://aclanthology.org/2021.emnlp-main.31} {Dynamic
  knowledge distillation for pre-trained language models}.
\newblock In \emph{Proceedings of the 2021 Conference on Empirical Methods in
  Natural Language Processing}, pages 379--389, Online and Punta Cana,
  Dominican Republic. Association for Computational Linguistics.

\bibitem[{Liang et~al.(2021)Liang, Hao, Shen, Zhou, Chen, Chen, and
  Carin}]{liang2020mixkd}
Kevin~J. Liang, Weituo Hao, Dinghan Shen, Yufan Zhou, Weizhu Chen, Changyou
  Chen, and Lawrence Carin. 2021.
\newblock \href {https://openreview.net/forum?id=UFGEelJkLu5} {Mixkd: Towards
  efficient distillation of large-scale language models}.
\newblock In \emph{9th International Conference on Learning Representations,
  {ICLR} 2021, Virtual Event, Austria, May 3-7, 2021}. OpenReview.net.

\bibitem[{Liang et~al.(2020)Liang, Li, Zheng, Lim, Salakhutdinov, and
  Morency}]{liang2020towards}
Paul~Pu Liang, Irene~Mengze Li, Emily Zheng, Yao~Chong Lim, Ruslan
  Salakhutdinov, and Louis-Philippe Morency. 2020.
\newblock \href {https://doi.org/10.18653/v1/2020.acl-main.488} {Towards
  debiasing sentence representations}.
\newblock In \emph{Proceedings of the 58th Annual Meeting of the Association
  for Computational Linguistics}, pages 5502--5515, Online. Association for
  Computational Linguistics.

\bibitem[{Liu et~al.(2018)Liu, Huang, Gao, Wei, Tian, and
  Liu}]{liu-etal-2018-task}
Qian Liu, Heyan Huang, Yang Gao, Xiaochi Wei, Yuxin Tian, and Luyang Liu. 2018.
\newblock \href {https://aclanthology.org/C18-1172} {Task-oriented word
  embedding for text classification}.
\newblock In \emph{Proceedings of the 27th International Conference on
  Computational Linguistics (COLING)}, Santa Fe, New Mexico, USA. Association
  for Computational Linguistics.

\bibitem[{Liu et~al.(2019)Liu, Ott, Goyal, Du, Joshi, Chen, Levy, Lewis,
  Zettlemoyer, and Stoyanov}]{Liu2019RoBERTa}
Yinhan Liu, Myle Ott, Naman Goyal, Jingfei Du, Mandar Joshi, Danqi Chen, Omer
  Levy, Mike Lewis, Luke Zettlemoyer, and Veselin Stoyanov. 2019.
\newblock \href {https://arxiv.org/abs/1907.11692} {Ro{BERT}a: {A} robustly
  optimized {BERT} pretraining approach}.
\newblock \emph{ArXiv preprint}, abs/1907.11692.

\bibitem[{Luo et~al.(2019)Luo, Wang, Fang, Hu, Tao, and Song}]{Luo19CFL}
Sihui Luo, Xinchao Wang, Gongfan Fang, Yao Hu, Dapeng Tao, and Mingli Song.
  2019.
\newblock \href {https://doi.org/10.24963/ijcai.2019/428} {Knowledge
  amalgamation from heterogeneous networks by common feature learning}.
\newblock In \emph{Proceedings of the Twenty-Eighth International Joint
  Conference on Artificial Intelligence, {IJCAI} 2019, Macao, China, August
  10-16, 2019}, pages 3087--3093. ijcai.org.

\bibitem[{Phan et~al.(2008)Phan, Nguyen, and Horiguchi}]{phan08gs}
Xuan~Hieu Phan, Minh~Le Nguyen, and Susumu Horiguchi. 2008.
\newblock \href {https://doi.org/10.1145/1367497.1367510} {Learning to classify
  short and sparse text {\&} web with hidden topics from large-scale data
  collections}.
\newblock In \emph{Proceedings of the 17th International Conference on World
  Wide Web, {WWW} 2008, Beijing, China, April 21-25, 2008}, pages 91--100.
  {ACM}.

\bibitem[{Pradhan et~al.(2013)Pradhan, Moschitti, Xue, Ng, Bj{\"o}rkelund,
  Uryupina, Zhang, and Zhong}]{pradhan2013ontonotes}
Sameer Pradhan, Alessandro Moschitti, Nianwen Xue, Hwee~Tou Ng, Anders
  Bj{\"o}rkelund, Olga Uryupina, Yuchen Zhang, and Zhi Zhong. 2013.
\newblock Towards robust linguistic analysis using ontonotes.
\newblock In \emph{Proceedings of the Seventeenth Conference on Computational
  Natural Language Learning}, pages 143--152.

\bibitem[{Raffel et~al.(2020)Raffel, Shazeer, Roberts, Lee, Narang, Matena,
  Zhou, Li, and Liu}]{raffel20t5}
Colin Raffel, Noam Shazeer, Adam Roberts, Katherine Lee, Sharan Narang, Michael
  Matena, Yanqi Zhou, Wei Li, and Peter~J. Liu. 2020.
\newblock Exploring the limits of transfer learning with a unified text-to-text
  transformer.
\newblock \emph{JMLR}, 21:140:1--140:67.

\bibitem[{Romero et~al.(2015)Romero, Ballas, Kahou, Chassang, Gatta, and
  Bengio}]{romero15fitnet}
Adriana Romero, Nicolas Ballas, Samira~Ebrahimi Kahou, Antoine Chassang, Carlo
  Gatta, and Yoshua Bengio. 2015.
\newblock \href {http://arxiv.org/abs/1412.6550} {Fitnets: Hints for thin deep
  nets}.
\newblock In \emph{3rd International Conference on Learning Representations,
  {ICLR} 2015, San Diego, CA, USA, May 7-9, 2015, Conference Track
  Proceedings}.

\bibitem[{Sang and De~Meulder(2003)}]{sang2003conll}
Erik~F Sang and Fien De~Meulder. 2003.
\newblock Introduction to the conll-2003 shared task: Language-independent
  named entity recognition.
\newblock \emph{arXiv preprint cs/0306050}.

\bibitem[{Schick et~al.(2021)Schick, Udupa, and Schütze}]{schick2020self}
Timo Schick, Sahana Udupa, and Hinrich Schütze. 2021.
\newblock \href {https://arxiv.org/abs/2103.00453} {Self-diagnosis and
  self-debiasing: A proposal for reducing corpus-based bias in nlp}.
\newblock \emph{ArXiv preprint}, abs/2103.00453.

\bibitem[{Schwartz et~al.(2020)Schwartz, Dodge, Smith, and
  Etzioni}]{schwartz2020green}
Roy Schwartz, Jesse Dodge, Noah~A Smith, and Oren Etzioni. 2020.
\newblock Green ai.
\newblock \emph{Communications of the ACM}, 63(12):54--63.

\bibitem[{Shen et~al.(2019)Shen, Wang, Song, Sun, and
  Song}]{shen19comprehensive}
Chengchao Shen, Xinchao Wang, Jie Song, Li~Sun, and Mingli Song. 2019.
\newblock \href {https://doi.org/10.1609/aaai.v33i01.33013068} {Amalgamating
  knowledge towards comprehensive classification}.
\newblock In \emph{The Thirty-Third {AAAI} Conference on Artificial
  Intelligence, {AAAI} 2019, The Thirty-First Innovative Applications of
  Artificial Intelligence Conference, {IAAI} 2019, The Ninth {AAAI} Symposium
  on Educational Advances in Artificial Intelligence, {EAAI} 2019, Honolulu,
  Hawaii, USA, January 27 - February 1, 2019}, pages 3068--3075. {AAAI} Press.

\bibitem[{Srivastava et~al.(2014)Srivastava, Hinton, Krizhevsky, Sutskever, and
  Salakhutdinov}]{srivastava2014dropout}
Nitish Srivastava, Geoffrey Hinton, Alex Krizhevsky, Ilya Sutskever, and Ruslan
  Salakhutdinov. 2014.
\newblock Dropout: a simple way to prevent neural networks from overfitting.
\newblock \emph{JMLR}, 15(1):1929--1958.

\bibitem[{Strubell et~al.(2019)Strubell, Ganesh, and
  McCallum}]{strubell-etal-2019-energy}
Emma Strubell, Ananya Ganesh, and Andrew McCallum. 2019.
\newblock \href {https://doi.org/10.18653/v1/P19-1355} {Energy and policy
  considerations for deep learning in {NLP}}.
\newblock In \emph{Proceedings of the 57th Annual Meeting of the Association
  for Computational Linguistics}, pages 3645--3650, Florence, Italy.
  Association for Computational Linguistics.

\bibitem[{Sun et~al.(2016)Sun, Li, Guo, Yu, Zheng, Si, and Liu}]{sun2016thuctc}
Maosong Sun, Jingyang Li, Zhipeng Guo, Z~Yu, Y~Zheng, X~Si, and Z~Liu. 2016.
\newblock Thuctc: an efficient chinese text classifier.
\newblock \emph{GitHub Repository}.

\bibitem[{Sun et~al.(2019)Sun, Cheng, Gan, and Liu}]{Sun2019PatientKD}
Siqi Sun, Yu~Cheng, Zhe Gan, and Jingjing Liu. 2019.
\newblock \href {https://doi.org/10.18653/v1/D19-1441} {Patient knowledge
  distillation for {BERT} model compression}.
\newblock In \emph{Proceedings of the 2019 Conference on Empirical Methods in
  Natural Language Processing and the 9th International Joint Conference on
  Natural Language Processing (EMNLP-IJCNLP)}, pages 4323--4332, Hong Kong,
  China. Association for Computational Linguistics.

\bibitem[{Thadajarassiri et~al.(2021)Thadajarassiri, Hartvigsen, Kong, and
  Rundensteiner}]{Thadajarassiri21SKA}
Jidapa Thadajarassiri, Thomas Hartvigsen, Xiangnan Kong, and Elke~A.
  Rundensteiner. 2021.
\newblock Semi-supervised knowledge amalgamation for sequence classification.
\newblock In \emph{AAAI}, pages 9859--9867.

\bibitem[{Vongkulbhisal et~al.(2019)Vongkulbhisal, Vinayavekhin, and
  Scarzanella}]{Vongkulbhisal19UHC}
Jayakorn Vongkulbhisal, Phongtharin Vinayavekhin, and Marco~Visentini
  Scarzanella. 2019.
\newblock \href {https://doi.org/10.1109/CVPR.2019.00329} {Unifying
  heterogeneous classifiers with distillation}.
\newblock In \emph{{IEEE} Conference on Computer Vision and Pattern
  Recognition, {CVPR} 2019, Long Beach, CA, USA, June 16-20, 2019}, pages
  3175--3184. Computer Vision Foundation / {IEEE}.

\bibitem[{Wang et~al.(2020)Wang, Wei, Dong, Bao, Yang, and
  Zhou}]{wang2020minilm}
Wenhui Wang, Furu Wei, Li~Dong, Hangbo Bao, Nan Yang, and Ming Zhou. 2020.
\newblock \href
  {https://proceedings.neurips.cc/paper/2020/hash/3f5ee243547dee91fbd053c1c4a845aa-Abstract.html}
  {Minilm: Deep self-attention distillation for task-agnostic compression of
  pre-trained transformers}.
\newblock In \emph{Advances in Neural Information Processing Systems 33: Annual
  Conference on Neural Information Processing Systems 2020, NeurIPS 2020,
  December 6-12, 2020, virtual}.

\bibitem[{Wolf et~al.(2020)Wolf, Debut, Sanh, Chaumond, Delangue, Moi, Cistac,
  Rault, Louf, Funtowicz, Davison, Shleifer, von Platen, Ma, Jernite, Plu, Xu,
  Le~Scao, Gugger, Drame, Lhoest, and Rush}]{wolf-etal-2020-transformers}
Thomas Wolf, Lysandre Debut, Victor Sanh, Julien Chaumond, Clement Delangue,
  Anthony Moi, Pierric Cistac, Tim Rault, Remi Louf, Morgan Funtowicz, Joe
  Davison, Sam Shleifer, Patrick von Platen, Clara Ma, Yacine Jernite, Julien
  Plu, Canwen Xu, Teven Le~Scao, Sylvain Gugger, Mariama Drame, Quentin Lhoest,
  and Alexander Rush. 2020.
\newblock \href {https://doi.org/10.18653/v1/2020.emnlp-demos.6} {Transformers:
  State-of-the-art natural language processing}.
\newblock In \emph{Proceedings of the 2020 Conference on Empirical Methods in
  Natural Language Processing: System Demonstrations}, pages 38--45, Online.
  Association for Computational Linguistics.

\bibitem[{Wu et~al.(2021)Wu, Wu, and Huang}]{wu-etal-2021-one}
Chuhan Wu, Fangzhao Wu, and Yongfeng Huang. 2021.
\newblock \href {https://doi.org/10.18653/v1/2021.findings-acl.387} {One
  teacher is enough? pre-trained language model distillation from multiple
  teachers}.
\newblock In \emph{Findings of the Association for Computational Linguistics:
  ACL-IJCNLP 2021}, pages 4408--4413, Online. Association for Computational
  Linguistics.

\bibitem[{Ye et~al.(2019)Ye, Ji, Wang, Ou, Tao, and Song}]{Ye19Master}
Jingwen Ye, Yixin Ji, Xinchao Wang, Kairi Ou, Dapeng Tao, and Mingli Song.
  2019.
\newblock \href {https://doi.org/10.1109/CVPR.2019.00294} {Student becoming the
  master: Knowledge amalgamation for joint scene parsing, depth estimation, and
  more}.
\newblock In \emph{{IEEE} Conference on Computer Vision and Pattern
  Recognition, {CVPR} 2019, Long Beach, CA, USA, June 16-20, 2019}, pages
  2829--2838. Computer Vision Foundation / {IEEE}.

\bibitem[{Zhang et~al.(2015)Zhang, Zhao, and LeCun}]{cnn15zhang}
Xiang Zhang, Junbo~Jake Zhao, and Yann LeCun. 2015.
\newblock \href
  {https://proceedings.neurips.cc/paper/2015/hash/250cf8b51c773f3f8dc8b4be867a9a02-Abstract.html}
  {Character-level convolutional networks for text classification}.
\newblock In \emph{Advances in Neural Information Processing Systems 28: Annual
  Conference on Neural Information Processing Systems 2015, December 7-12,
  2015, Montreal, Quebec, Canada}, pages 649--657.

\bibitem[{Zmigrod et~al.(2019)Zmigrod, Mielke, Wallach, and
  Cotterell}]{zmigrod2019cda}
Ran Zmigrod, Sabrina~J. Mielke, Hanna Wallach, and Ryan Cotterell. 2019.
\newblock \href {https://doi.org/10.18653/v1/P19-1161} {Counterfactual data
  augmentation for mitigating gender stereotypes in languages with rich
  morphology}.
\newblock In \emph{Proceedings of the 57th Annual Meeting of the Association
  for Computational Linguistics}, pages 1651--1661, Florence, Italy.
  Association for Computational Linguistics.

\end{thebibliography}
\bibliographystyle{acl_natbib}

\appendix
\section{Datasets Details}
\label{apx:dataset}
The label sets of datasets we used in the main paper are first sorted according to the name and will be evenly divided into subsets according to the number of teacher models. 
Table~\ref{tab:dataset} gives the dataset statistics and the class number for two teacher models experiments.
Table~\ref{tab:class_split} gives the label~(or entity types for NER datasets) list on each dataset.
For teacher number that cannot evenly dividing the label sets, the final label set will include the left labels.
For example, when there are $4$ teacher models are needed on THUCNews dataset, the label set will be split into  \{\emph{Sports}, \emph{Enter.}\}, \{\emph{Furniture}, \emph{Estate}\},  \{\emph{Education}, \emph{Fashion}\} and \{\emph{Politics}, \emph{Game},\emph{Tech.}, \emph{Finance}\}.
% % class split details for training teacher models in different settings.
% Specifically, we divide the original dataset into two parts with disjoint classes. 
% The specialty class are listed in Table~\ref{tab:class_split}

\begin{table}[ht!]
    \centering
    \small 
    \begin{tabular}{@{}l|crrc@{}}
    \toprule
      Dataset  & {\#Class~(Ent.)} &  {\#Train} & {\#Test} &  $\{|Y|\}$ \\
      \midrule 
      AG News  & 4  & 120k & 7.6k & \{2, 2\}\\ 
    %   DBPedia & 14 & 560k& 70k\\ 
      5Abstracts Group & 5  & 53k& 1k  & \{2, 3\} \\
    %   Yahoo Answers & 10& 1,400k& 60k \\d
    Google Snippets & 8 & 10k & 2.2k& \{4, 4\} \\ 
    %   Sogou News &  6 & 60k & 6k \\ 
    %   20News Group &  20 & 20k & \\
      THUCNews & 10  & 50k & 10k&  \{5, 5\}\\
    %   TNews & 15 &  53k & 10k \\ 
    \midrule 
    CoNLL 2003 & 4 & 14k & 3.5k& \{2, 2\}\\ 
    Ontonotes v5 & 18 & 115.8k &12.2k & \{9, 9\}\\ 	
    \bottomrule
    \end{tabular}
    \caption{Statistics of datasets used in our paper. Ent. denotes the entity types and $\{|Y|\}$ is the number of classes each teacher model specializes.}
    \label{tab:dataset}
\end{table}

\begin{table}[tb!]
    \centering
    \small 
    \begin{tabular}{@{}l| ll@{}}
    \toprule
      Dataset   & Label Order  \\
    \midrule 
     AG News    & World, Sports,  Business, Sci/Tech  \\ 
     \midrule 
     \multirow{3}{*}{THUCNews}  & Sports, Entertainment, Furniture, \\
     & Estate, Education, Fashion, \\
     & Politics, Game, Technology, Finance\\
      \midrule 
     \multirow{4}{*}{Google Snippets} &   Business, Computers, \\ 
     & Culture-Arts-Entertainment, \\
     & Education-Science, Engineering, \\
     & Health, Politics-Society, Sports \\
     \midrule
     5Abstracts Group & Business, CSAI, Law,  Sociology, Trans\\ 
          \midrule
     CoNLL 2003 & PER, MISC, ORG, LOC\\ 
          \midrule
     \multirow{6}{*}{OntoNotes 5.0} & LAW, TIME, DATE, LOC, \\
        & ORG, GPE, NORP, LANGUAGE, \\ 
        & EVENT, CARDINAL, MONEY, \\
        & PRODUCT, PERCENT, FAC, \\
        & PERSON, QUANTITY,\\
        & WORK OF ART, ORDINAL\\ 
\bottomrule
    \end{tabular}
    \caption{Sorted label names~(entity types) of datasets. Label names of THUCNews are translated into English.}
    \label{tab:class_split}
\end{table}

\section{Hyper-parameter Search for $\tau$}
\label{apx:temp}
\begin{figure}[tbh!]
    \centering
    \includegraphics[width=0.95\linewidth]{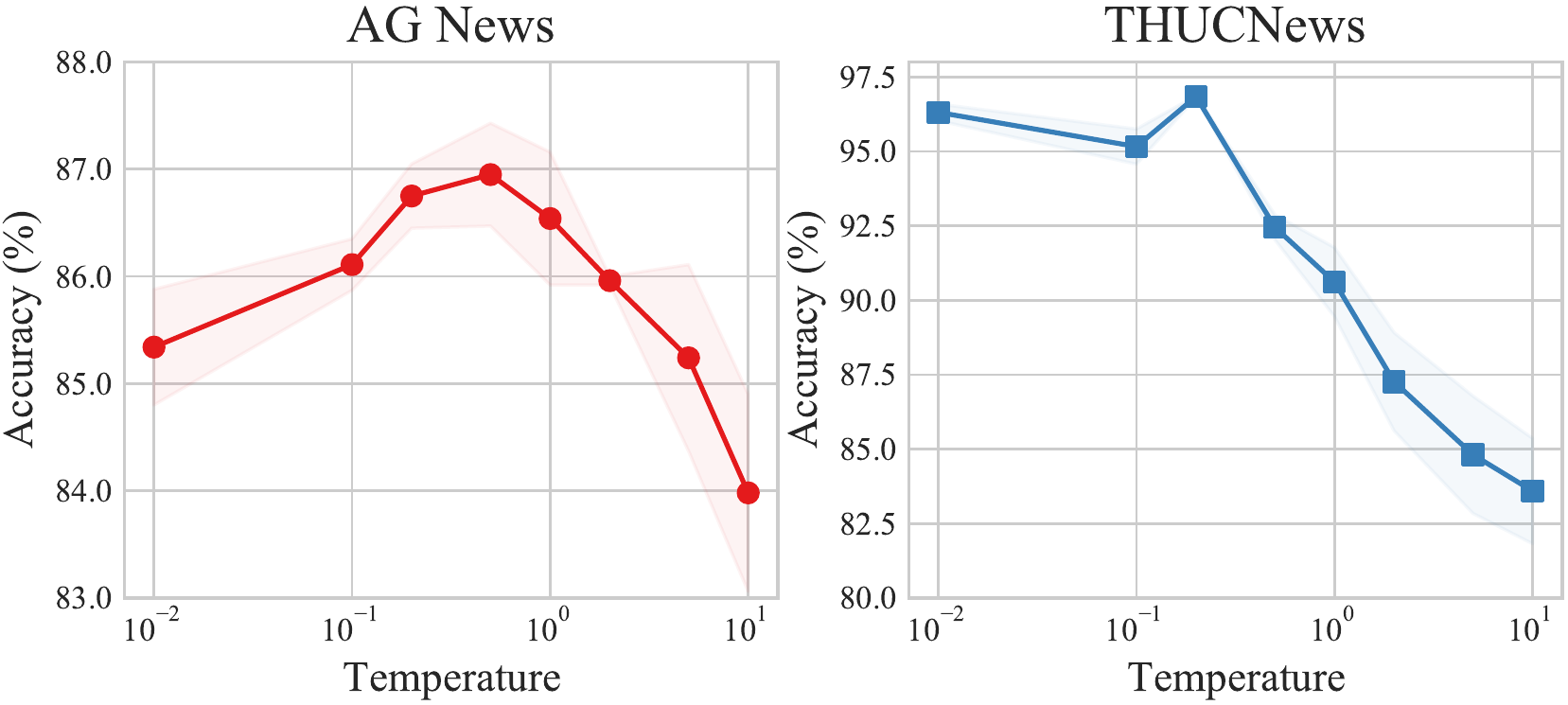}
    \caption{Varying temperature $\tau$ for MUKI-Soft. The average accuracy of three seeds are plotted with standard deviation in shade.}
    \label{fig:temperature_curve}
\end{figure}

% In MUKI-Soft, $\tau$ is a hyper-parameter controlling the smoothness of teacher supervision weights. 
We perform a hyper-parameter search experiment for the optimal $\tau$ in MUKI-soft.
% We vary the temperature $\tau$ for controlling the smoothness of supervision weights in MUKA-Soft. 
We conduct experiments on AG News and THUCNews for stable results.
The values of $\tau$ are picked from $\{0.01, 0.1, 0.2, 0.5, 1.0, 2.0, 5.0, 10.0\}$, and the results are shown in Figure~\ref{fig:temperature_curve}. 
We observe that the accuracy drops significantly when $\tau$ is set to high values, where the teacher weights distribution is becoming a uniform distribution, while it reaches a peak when $\tau$ is set to a small value between $0.2$ and $0.5$.
It indicates that slightly sharpening the teacher weights distribution is helpful for KI.
Therefore, we adopt $\tau = 0.2$ in all the experiments.

\end{document}